\documentclass[journal]{IEEEtran}

\usepackage[latin1]{inputenc}
\usepackage{amsmath}
\usepackage{graphicx}
\usepackage{listings}
\usepackage[colorlinks,linkcolor=black,citecolor=black,urlcolor=black]{hyperref}
\usepackage{siunitx}
\usepackage{cite}
\usepackage{fixltx2e}
\usepackage{multirow}
\usepackage{threeparttable}
\usepackage{tabularx}
\usepackage{booktabs}
\usepackage{units}
\usepackage{xcolor}
\usepackage[normalem]{ulem}
\usepackage{float}

\newif\ifshowToDos

\showToDosfalse

\newcommand{\red}[1]{\textcolor{red}{#1}}

\ifshowToDos
  \usepackage{todonotes}
  \paperwidth=\dimexpr \paperwidth + 8cm\relax
  \oddsidemargin=\dimexpr\oddsidemargin + 4cm\relax
  \evensidemargin=\dimexpr\evensidemargin + 4cm\relax
  \marginparwidth=\dimexpr \marginparwidth + 4cm\relax
\else
  \usepackage[disable]{todonotes}
\fi

\definecolor{gray}{gray}{0.2}

\title{Low-Power Low-Latency Keyword Spotting and Adaptive Control with a SpiNNaker 2 Prototype and Comparison with Loihi}

\author{Yexin Yan, Terrence C. Stewart, Xuan Choo, Bernhard Vogginger, \\ Johannes Partzsch, Sebastian H\"oppner, Florian Kelber, Chris Eliasmith, Steve Furber, Christian Mayr
\thanks{Manuscript received \today. The research leading to these results has received funding from the European Union (EU) Seventh Framework Programme (FP7) under grant agreement no 604102, the EU's Horizon 2020 research and innovation programme under grant agreements No 720270 and 785907 (Human Brain Project, HBP), Intel Labs, the Canada Research Chairs program, Natural Sciences and Engineering Research Council of Canada (NSERC) Discovery grant 261453, and the National Research Council Canada (NRCC) at the University of Waterloo. The authors thank Arm and Racyics GmbH for IP.}

\thanks{Y. Yan, B. Vogginger, J. Partzsch, S. H\"oppner, F. Kelber and C. Mayr are with the Faculty of Electrical and Computer Engineering, Technische Universit\"at Dresden, Germany (e-mail: yexin.yan@tu-dresden.de) }
\thanks{X. Choo and C. Eliasmith are with Applied Brain Research, Inc, Waterloo, Canada}
\thanks{T.C. Stewart is with the National Research Council of Canada, University of Waterloo Collaboration Centre, Waterloo, Canada}
\thanks{S. Furber is with the Advanced Processor Technologies Research Group, University of Manchester, UK}
}

\begin{document}
\bstctlcite{IEEEexample:BSTcontrol}

\maketitle

\begin{abstract}
  We implemented two neural network based benchmark tasks on a prototype chip of the second-generation SpiNNaker (SpiNNaker 2) neuromorphic system: keyword spotting and adaptive robotic control. Keyword spotting is commonly used in smart speakers to listen for wake words, and adaptive control is used in robotic applications to adapt to unknown dynamics in an online fashion. We highlight the benefit of a multiply-accumulate (MAC) array in the SpiNNaker 2 prototype which is ordinarily used in rate-based machine learning networks when employed in a neuromorphic, spiking context. In addition, the same benchmark tasks have been implemented on the Loihi neuromorphic chip, giving a side-by-side comparison regarding power consumption and computation time. While Loihi shows better efficiency when less complicated vector-matrix multiplication is involved, with the MAC array, the SpiNNaker 2 prototype shows better efficiency when high dimensional vector-matrix multiplication is involved.
\end{abstract}

\begin{IEEEkeywords}
  SpiNNaker, MAC array, Loihi, neuromorphic computing
\end{IEEEkeywords}

\section{Introduction}
\label{sec:intro}

With the substantial progress of artificial intelligence (AI) in recent years, neural network based algorithms are increasingly being deployed in embedded AI applications. Smart speakers which continuously listen for keywords like "Alexa" and robotic applications which employ neural network based adaptive control algorithms are examples from industry and research. To improve the efficiency regarding power consumption and computation time, various hardware architectures have been proposed. 

The neural networks employed in these AI applications are most commonly deep neural networks (DNNs). A substantial amount of computation in DNNs is caused by the multiply-accumulate (MAC) operations. For efficient computation of DNNs, many machine learning hardware architectures include a MAC unit to facilitate the MAC operations in DNNs \cite{mlacc_survey2017}. 

While DNNs are currently widely adopted for applications, spiking neural networks (SNNs) which more closely mimic the behavior of biological neural networks are increasingly gaining attention as this type of network has the potential of high efficiency, especially in combination with neuromorphic hardware \cite{Roy2019}. One prominent example is the Loihi neuromorphic chip \cite{loihi} with dedicated circuits for synapse and neuron models and a programmable learning engine, which has been shown to be efficient in various neural network based benchmark tasks like keyword spotting \cite{keyword_spotting_loihi19} and adaptive control \cite{nengo_adaptive_loihi2020}. Another neuromorphic architecture is represented by the second generation of the SpiNNaker system (SpiNNaker 2) \cite{spinn2arxiv} with general purpose processors connected with numerical  accelerators. Besides neuromorphic accelerators, SpiNNaker2 also contains MAC arrays and is thus able to merge SNN and DNN operation. 

In this work, we implement the keyword spotting and adaptive control benchmark tasks on the second SpiNNaker 2 prototype \red{(cite jib1 paper)}. We compare the computation time and active energy consumption of the benchmark tasks with Loihi, and highlight the benefit of the MAC array. Specifically, for keyword spotting, because the original DNN version is implemented on the SpiNNaker 2 prototype with the MAC array, and the SNN version is implemented on Loihi because it only supports SNN, the SpiNNaker 2 prototype shows better efficiency regarding computation time and energy consumption. For adaptive control, SNN is implemented on both hardwares and Loihi shows better efficiency when low dimensional vector-matrix multiplication is involved, and the SpiNNaker 2 prototype shows better efficiency when high dimensional vector-matrix multiplication is involved.

In Section \ref{sec:hardware} we give an overview of the prototype chip, with emphasis on the MAC array. Section \ref{sec:benchmarks} describes the two benchmarks implemented in this work. Section \ref{sec:software} presents the software implementation. The experimental results are presented in Section \ref{sec:results}.

\section{The SpiNNaker 2 Prototype Chip}
\label{sec:hardware}
\subsection{System Overview}\label{sys_ove}
SpiNNaker \cite{furber14} is a digital neuromorphic hardware system based on low-power Arm processors originally built for  real-time simulation of spiking neural networks (SNNs). In the second generation of SpiNNaker (SpiNNaker 2), which is currently being developed in the Human Brain Project \cite{amunts2016human}, several improvements are being made. The SpiNNaker 2 architecture is based on Processing Elements (PEs) which contain an Arm Cortex-M4F core, 128 KBytes local SRAM, hardware accelerators for exponential functions \cite{partzsch17} and true- and pseudo random numbers \cite{felix16,synsampling19} and multiply-accumulate (MAC) accelerators. Additionally, the PEs include advanced dynamic voltage and frequency scaling (DVFS) features \cite{hoeppner17iscas,hoeppner19tcas}. The PEs are arranged in Quad-Processing Elements (QPEs) containing four PEs and a Network-on-Chip (NoC) router for packet based on-chip communication. The QPEs can be placed in a array scheme without any additional flat toplevel routing to form the SpiNNaker 2 many core SoC.

SpiNNaker 2 will be implemented in GLOBALFOUNDRIES 22FDX technology ~\cite{glofo16}. This FDSOI technology allows the application of adaptive body biasing (ABB) for low-power operation at ultra-low supply voltages in both forward \cite{Hoeppner2019a} and reverse bias schemes \cite{Walter2020}. For maximum energy efficiency and reasonable clock frequencies, 0.50V nominal supply voltage is chosen and ABB in a forward bias scheme is applied. The ABB aware implementation methodology from \cite{Hoeppner2019b} has been used. This allows to achieve \SI{>200}{\MHz} clock frequency at 0.50V nominal supply voltage at the first DVFS performance level PL1 and \SI{>400}{\MHz} from 0.60V supply at the second DVFS performance level PL2.  

\begin{figure}[htb]
  \centering
  \includegraphics[width=0.48\textwidth]{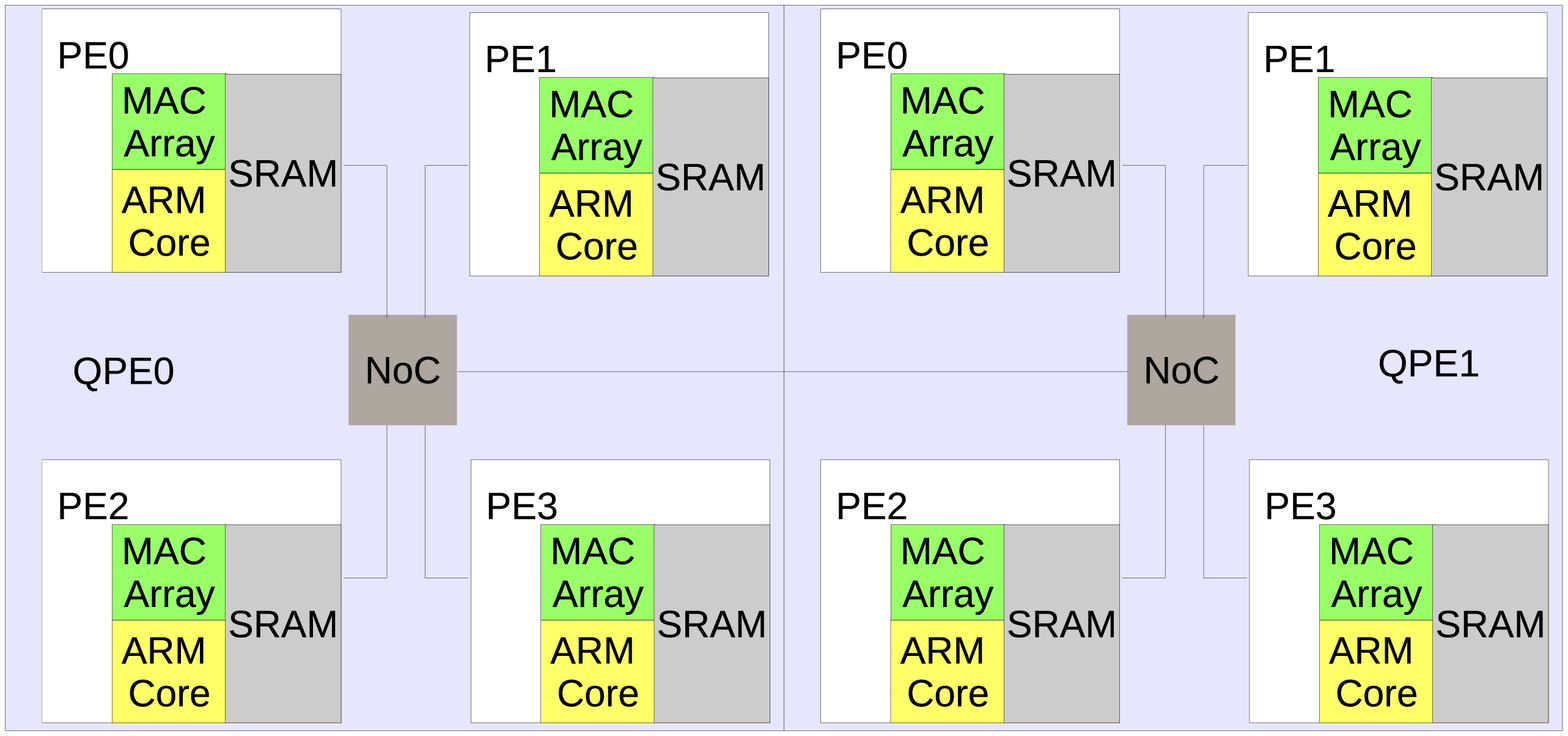}
  \caption{Simplified schematic of the second SpiNNaker 2 prototype with 2 QPEs. Each QPE contains 4 PEs. Each PE contains a MAC array, an Arm core and a local SRAM. The NoC router is responsible for the communication. 
  }\label{fig:pe_schematic} 
\end{figure}

\begin{figure}[htb]
  \centering
  \includegraphics[width=0.25\textwidth]{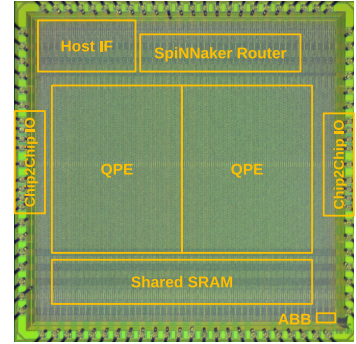}
  \caption{Chip photo of the SpiNNaker 2 prototype in 22FDX technology 
  }\label{fig:chipphoto} 
\end{figure}

The second SpiNNaker 2 prototype chip has been implemented and manufactured in 22FDX \textcolor{red}{(cite Jib1 paper)}. It contains 2 QPEs with 8 PEs in total to allow the execution of neuromorphic applications. Fig.~\ref{fig:pe_schematic} shows the simplified block diagram of the testchip PE array. The chip photo is shown in Fig.~\ref{fig:chipphoto}. The testchip includes peripheral components for host communication, a prototype of the SpiNNaker router for chip-to-chip spike communication and some shared on-chip SRAM.   

\subsection{MAC Array}\label{mac_acc}

\begin{figure}[htb]
  \centering
  \includegraphics[width=0.4\textwidth]{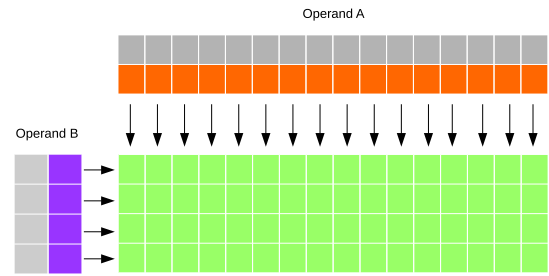}
  \caption{Schematic of the MAC array. Each square in the 4 x 16 block represents one MAC unit. The squares around the block represent the data to be executed. In each clock cycle, 4 values from operand B and 16 values from operand A are fed into the MAC array simultaneously, as indicated by the arrows. 
  }\label{fig:mac_schematic} 
\end{figure}

The MAC array has 64 MAC units in a 4 x 16 layout. Fig. \ref{fig:mac_schematic} illustrates the MAC array. The data of operand A and operand B are arrays of 8 bit integer values. In each clock cycle, 16 values from the array of operand A and 4 values from the array of operand B are fed into the MAC array. Every MAC unit in the same column is fed with the same value from operand A, and every MAC unit in the same row is fed with the same value from operand B. The software running on the Arm core is responsible for arranging the data in the SRAM and notifying the MAC array the address and length of the data to be processed. After the data is processed, the results are written back to predefined addresses in the memory. The result of each MAC unit is 29-bit.

When computing a matrix multiplication, a general purpose processor like the Arm core needs to: 1. fetch the operand A and operand B into the registers, 2. do the multiply-accumulate, 3. write the result back, 4. check the condition of the loop, 5. compute the addresses of the data in the next iteration. While the MAC array essentially does the same, it is more efficient due to the Single Instruction Multiple Data (SIMD) operation. In particular, the efficiency is made possible by:

1. 64 MAC operations can be done in one clock cycle in parallel.

2. 16 x 8 bits of data of operand A and 4 x 8 bits of data of operand B can be fetched in one clock cycle in parallel

3. control logic and data transfer in parallel to MAC operations, hiding the overhead of data transfer for the next iteration. 


\section{Benchmark Models}
\label{sec:benchmarks}
In this section, we briefly review the two benchmark models implemented in this work: keyword spotting and adaptive control.   

\subsection{Keyword Spotting}\label{keyword_spotting}

\begin{figure}[htb]
  \centering
  \includegraphics[width=0.4\textwidth]{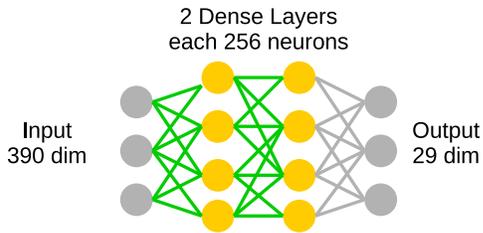}
  \caption{Keyword Spotting Network Architecture
  }\label{fig:keyword_spotting_algo} 
\end{figure}

Keyword spotting is a speech processing problem which deals with identifying keywords in utterances. A practical use case is the identification of wake words for virtual assistants (e.g. "Alexa"). In this work, the keyword spotting network we implement on the SpiNNaker 2 prototype is the same as in \cite{keyword_spotting_loihi19}, which consists of 1 input layer with 390 input values, 2 dense layers each with 256 neurons and 1 output layer with 29 output values (Fig. \ref{fig:keyword_spotting_algo}). Also, the same as in \cite{keyword_spotting_loihi19}, no training is involved and only inference is considered. The 390 dimensional input to the network is the Mel-frequency cepstral coefficient (MFCC) features of an audio waveform in each time step. The 29 dimensional output of the network basically corresponds to the alphabetical characters, with additional special characters for e.g. silence etc. One 'inference' with this network involves passing 10 time steps of the MFCC features into the network. The outputs are then postprocessed to form a result for the inference. The difference to the implementation on Loihi is that on the SpiNNaker 2 prototype, we implement the network with normal DNN with ReLU activations, whereas on Loihi, the SNN version was implemented since Loihi only supports SNNs. 

\subsection{Adaptive Control}\label{adaptive_control}

For our second benchmark task, we use the adaptive control algorithm proposed as a benchmark in \cite{terry15} and further investigated in \cite{nengo_adaptive_loihi2020}.  This benchmark consists of a single-hidden-layer neural network, where the input is the sensory state of the system to be controlled (such as a robot arm) and the output is the extra force that should be applied to compensate for the intrinsic dynamics and forces on the arm (gravity, friction, etc.)  The only non-linearities are in the hidden layer (i.e. there is no non-linear operation directly on the input or output).  The input weights are fixed and randomly chosen, and the output weights $\omega_{ij}$ are initialized to zero and then adjusted using a variant of the delta rule \cite{eliasmith2011} (Eq. \ref{eq:adaptive_control}), where $\alpha$ is a learning rate, $a_i$ is the current level of activity of the $i$th neuron, and $E_j$ is an error signal.
\begin{align}
\Delta \omega_{ij} = \alpha a_i E_j
\label{eq:adaptive_control}
\end{align}

Crucially, if we use the output of a PD-controller to be this error signal $E_j$, and if we take the output of this network and add it to the control signal produced by a PD-controller, then the resulting system will act as a stable adaptive controller \cite{dewolf16}.  This is a variant of the adaptive control algorithm developed by Jean-Jacques Slotine \cite{slotine87}.  One way to think of this is that the neural network is acting somewhat like the I term in a PID-controller, but since the I value is being produced by the neural network, it can be different for different parts of the sensory space.  It can thus learn to, for example, apply extra positive torque when a robot arm is leaning far to one side, and extra negative torque when the arm is leaning far to the other side.

When used with spiking neurons, we also apply a low-pass filter to the $a_i$ term, producing a continuous value representative of the recent spiking activity of the neuron.

While this benchmark was originally proposed for its simplicity and applicability across a wide range of neuromorphic hardware and controlled devices, there is one further important reason for us to choose this benchmark.  The core network that it requires has a single hidden layer non-linearity, and the inputs and outputs are generally of much lower dimensionality than the number of neurons in the hidden layer.  This is exactly the sort of network that forms the core component of the Neural Engineering Framework (NEF) \cite{eliasmith2003a}.  The NEF has been used to create large-scale biologically-based neural models \cite{eliasmith2012} by chaining these smaller networks together.  By sending the output from one of these networks to the inputs of another network, we are effectively factoring the weight matrix between the hidden layers of the two networks.  This has been shown to be a highly efficient method for implementing neural models on the original SpiNNaker 1 hardware \cite{mundy15}, and we expect the same to be the case on SpiNNaker 2.

\begin{figure}[htb]
  \centering
  \includegraphics[width=0.48\textwidth]{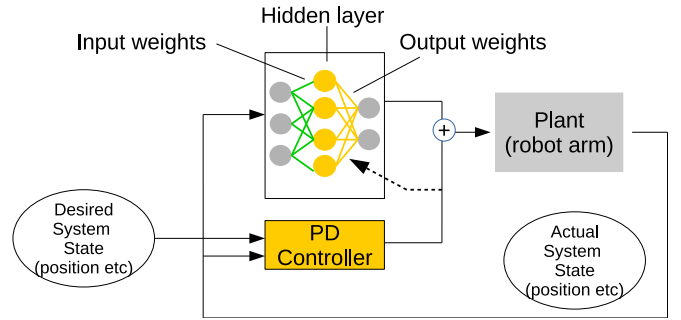}
  \caption{Adaptive Control Network Architecture
  }\label{fig:adaptive_algo} 
\end{figure}

\section{Implementation of the Benchmarks on the SpiNNaker 2 Prototype}
\label{sec:software}
We implemented the keyword spotting and adaptive control benchmarks on the SpiNNaker 2 prototype with the MAC array and ARM core responsible for different computational tasks. Since the same benchmarks have also been implemented on Loihi \cite{keyword_spotting_loihi19}\cite{nengo_adaptive_loihi2020}, this allows a side-by-side comparison between both neuromorphic hardwares.

\subsection{Keyword Spotting}

\begin{figure}[htb]
  \centering
  \includegraphics[width=0.48\textwidth]{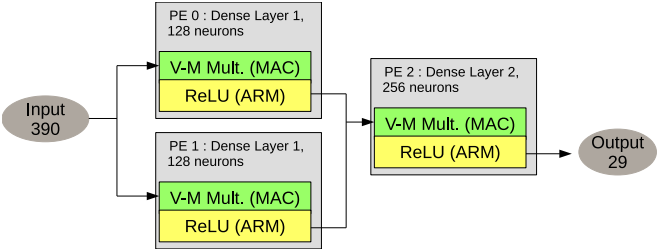}
  \caption{Implementation of keyword spotting network on the SpiNNaker 2 prototype
  }\label{fig:keywordspotting_impl} 
\end{figure}

The keyword spotting network consists of 2 computational steps: vector-matrix multiplication which is done with the MAC array and ReLU update which is done with the ARM core. Because of memory constraints (see Section \ref{result_keyword_spotting_memory}) layer 1 is split into 2 PEs. The weights in this network are the same as in \cite{keyword_spotting_loihi19}. The input to the network is a 390 dimensional vector of 8 bit integers. The ReLU activations of each layer are also 8 bit integers. The ReLU activations of layer 2 are directly sent back to host PC, where the vector-matrix multiplication for the output layer with 29 dimensions is performed, the same as in \cite{keyword_spotting_loihi19}. Fig. \ref{fig:keywordspotting_impl} shows the implementation of the keyword spotting network on the SpiNNaker 2 prototype. 

\subsection{Adaptive Control}

The implementation of adaptive control on the SpiNNaker 2 prototype is based on \cite{mundy15} and \cite{knight2016}. There are mainly 4 computational steps: input processing, neuron update, output processing and weight update. 

In input processing, the inputs to the network are multiplied with the input weight matrix to produce the input current for each neuron in the hidden layer. The weights are quantized to 8 bit integers with stochastic rounding. The vector-matrix multiplication with only ARM core and without MAC array is also implemented and serves as reference.

The rest of the computation is implemented on the ARM core which allows event based processing.

In neuron update, the neuron dynamics is updated according to the input current. The Leaky-Integrate-and-Fire (LIF) neuron model is used in the hidden layer to allow for event based processing of the spikes in the following steps.

In output processing, the outputs of the neurons are multiplied with the output weight matrix. In the case of non-spiking neuron models like ReLU, this process is a vector-matrix multiplication. In the case of spiking neuron models, a connection is only activated when there is a spike, so this output processing step corresponds to adding the weights associated with the neuron which has spiked to the output of the network. 

In weight update, the output weight matrix is updated according to the neuron activity and error signal. In order to do weight update in an event based manner, the low pass filter mentioned in section \ref{adaptive_control} has been removed, similar to \cite{knight2016}. Because of the short time constant of the low pass filter in this application, this modification doesn't affect the performance. Since the learning rate is normally very small, floating point data type is chosen for the weights in the output weight matrix.

In this work, we focus on the adaptive control network implemented on a single PE. The implementation is done with scalability in mind. In the case that the size of a neuron population exceeds the memory limit of a PE, it can be split into many PEs \cite{mundy15}. In this work, the PE additionally simulates the PD controller. The overhead is negligible.

\begin{figure}[htb]
  \centering
  \includegraphics[width=0.48\textwidth]{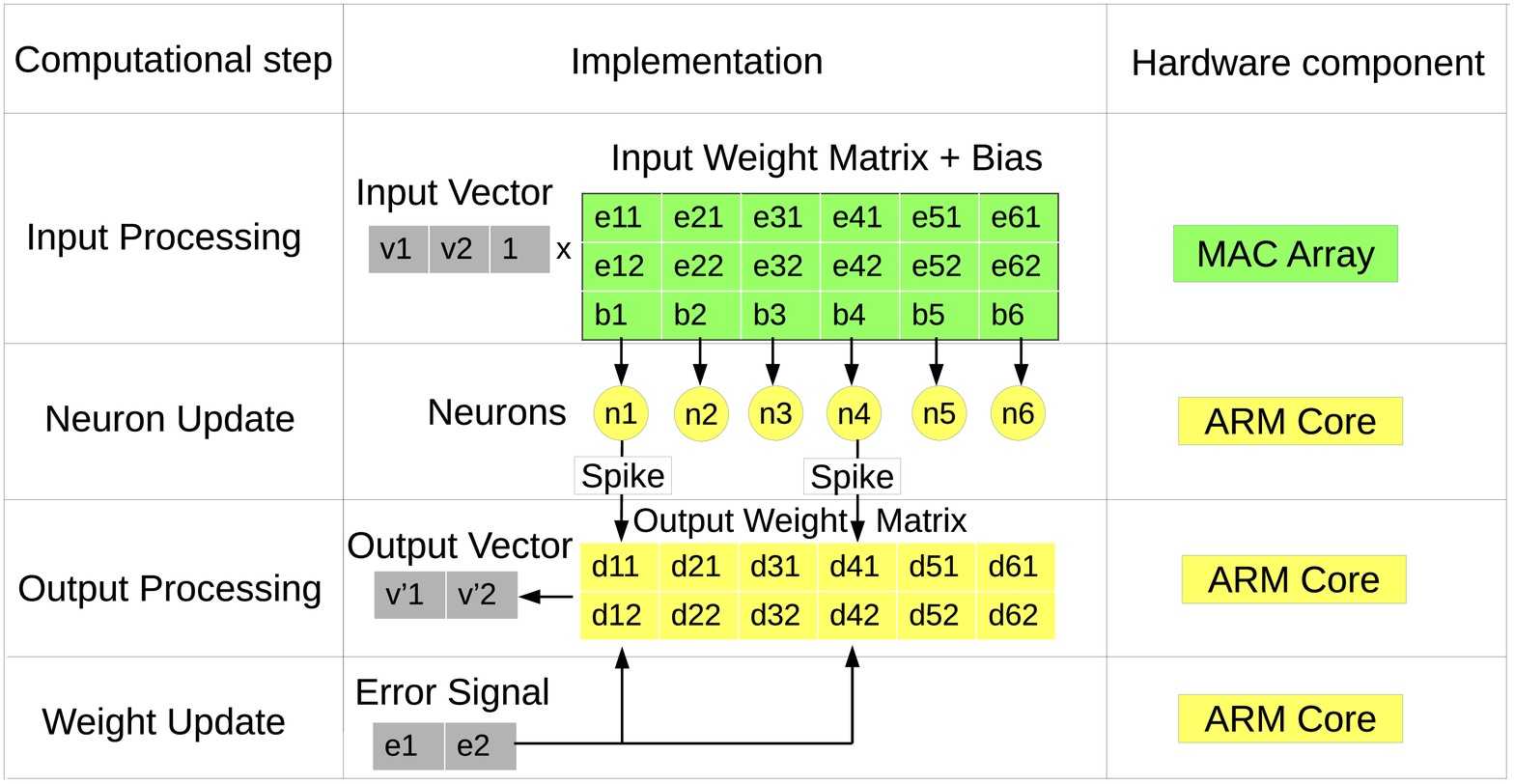}
  \caption{Main computational steps and hardware component for each step in adaptive control
  }\label{fig:nef_mlacc} 
\end{figure}

The computational steps and the hardware component used for each step is summarized in Fig.  \ref{fig:nef_mlacc}. The PD controller is not shown since the computation is relatively simple.

\section{Results}
\label{sec:results}

In this section we show the results of both benchmarks running on the SpiNNaker 2 prototype chip. In particular, we show results regarding the memory footprint, computation time and energy measurement when the PE is running with 0.5 V and 250 MHz. The results of computation time and energy measurement are compared with Loihi. In addition, for adaptive control, the SpiNNaker 2 prototype chip is connected to a robotic arm to demonstrate real time control. 
Since we implemented the same models on the SpiNNaker 2 prototype as on Loihi, the differences between both hardwares in terms of classification accuracy in the case of keyword spotting and mean squared error between actual and desired trajectories in the case of adaptive control are negligibly small, so that this will not be further discussed in this section. 

\subsection{Keyword Spotting}\label{result_keyword_spotting}

\subsubsection{Memory Footprint}\label{result_keyword_spotting_memory}

For the keyword spotting benchmark, the required SRAM memory mainly consists of 2 parts: weight memory and neuron input memory.

The weight memory is the memory for storing the weights and biases, which are quantized as 8-bit integers. The required memory in bytes is
\begin{equation}
M_{w} \;=\; (D + 1) N
\label{eq:M_w}
\end{equation}

where \(D\) is the number of input dimensions, \(N\) is the number of neurons. 

The neuron input memory is the memory for storing the results from the MAC array after the vector-matrix multiplication is complete. Each input is a 32 bit integer. The required memory in bytes is
\begin{equation}
M_{i} \;=\; 4N
\label{eq:M_ic_kw}
\end{equation}

Since the ReLU unit doesn't need to hold its output value between inferences, which is the case for the LIF neuron model, there is no neuron memory needed.

The total memory for a neural network on a PE is
\begin{equation}
M_{total} \;=\; M_{w}+M_{i}
\label{eq:M_total_kw}
\end{equation}

Based on Eq. (\ref{eq:M_w}), (\ref{eq:M_ic_kw}), (\ref{eq:M_total_kw}) for memory footprint, the first hidden layer of the keyword spotting network would require ca. 100 KBytes of memory. For each PE, in total 128 KBytes of SRAM memory is available, which is used for the program code as well as the program data. In this work, it is assumed that each PE has 90 KBytes of SRAM memory available for the data of the neural network. So the first hidden layer is split into two PEs. 

\subsubsection{Computation Time and Comparison with Loihi}\label{result_keyword_spotting_time}

In the keyword spotting benchmark, the computation times for the vector-matrix multiplication (\(T_{mm}\)) and the ReLU update (\(T_{relu}\)) are measured. After the measurement, polynomial models can be fitted by minimizing the mean-squared error. The number of clock cycles for the vector-matrix multiplication with the MAC array is found to be
\begin{align}
T_{mm} \;=\; 74.0  + 5.38 N \nonumber \\
+ 0.13 N D + 24.0 D
\label{eq:c_mm}
\end{align}

where \(N\) is the number of neurons and \(D\) is the number of input dimensions. The time for the vector-matrix multiplication is mostly reflected in \(0.13 N D\). Before the vector-matrix multiplication starts, the inputs to the network needs to be prepared for the MAC array. This pre-processing step is mostly reflected in \(24.0 D\). After the vector-matrix multiplication, a post-processing step is necessary for the resulting neuron input current. The computation time depends on both \(D\) and \(N\), and this is reflected in \(24.0 D\) and \(5.38 N\). For each of the computational steps, there is a constant overhead, which is reflected in the constant \(74.0\).

The computation time for ReLU update with ARM core is found to be
\begin{align}
T_{relu} \;=\; 17.70 N + 117.5
\label{eq:c_relu_neuron_update}
\end{align}

The total time is
\begin{align}
T_{total} \;=\; T_{mm} + T_{relu}
\label{eq:c_kw_total}
\end{align}

Based on Eq. (\ref{eq:c_mm}), (\ref{eq:c_relu_neuron_update}), (\ref{eq:c_kw_total}) for computation time, with the keyword spotting network split into 3 PEs (Fig. \ref{fig:keywordspotting_impl}), the computation of one time step consumes less than 21k clock cycles. With a safety margin of 4k clock cycles, one time step would take less than 25k clock cycles. When the PE is running with 250 MHz, this means the duration of one time step can be reduced to 0.1 ms. Since 10 time steps are combined to 1 time window to form one inference, a time step duration of 0.1 ms would correspond to 1000 inferences per second. In \cite{keyword_spotting_loihi19}, 296 inferences per second has been reported for Loihi. One reason for the reduced speed of Loihi might be that the inputs to the neural network are coming from an FPGA which could cause some latency, while the SpiNNaker 2 prototype is using inputs generated by one of the PEs of the same chip.


\subsubsection{Energy Measurement and Comparison with Loihi}\label{result_keyword_spotting_power}

Both QPEs are used for the measurement. In each QPE, 3 PEs are switched on to simulate a keyword spotting network. The measured result is then divided by 2 to obtain the energy per network. The energy is measured incrementally, similar to previous measurements on SpiNNaker 1 \cite{spinn1power} and on the first SpiNNaker 2 prototype \cite{hoeppner19tcas}. The idle energy is measured when in each time step the timer tick interrupt is handled but nothing is processed. The result we present in this section is the active energy which is obtained by subtracting the idle energy from the total energy. The resulting active energy per inference is 7.1 \(\mu\)J.

The keyword spotting network is implemented as a normal DNN on the SpiNNaker 2 prototype. The MAC array is used for the computation of the connection matrix, and the ARM core is used for the computation of ReLU activation function. Since Loihi only supports SNN, the spiking version of the keyword spotting network is implemented on Loihi. This could be the reason that the SpiNNaker 2 prototype consumes less energy for each inference in the keyword spotting benchmark (Tab. \ref{tab:keyword_spotting}). Note that in \cite{keyword_spotting_loihi19}, the reported energy per inference on Loihi was 270 \(\mu\)J, including a 70 mW overhead presumably caused by the x86 processor on Loihi. In this work the overhead has been removed which results in 37 \(\mu\)J per inference.

\begin{table}[htb]
  \begin{minipage}{0.47\textwidth}
    \centering
    \renewcommand{\arraystretch}{1.1}
    \caption{Comparison of the SpiNNaker 2 prototype (SpiNN) and Loihi for the keyword spotting task }
    \label{tab:keyword_spotting}
    \centering
    \footnotesize
    \begin{tabular}{lcc} \toprule
      Hardware    & inference/sec    & energy/inference (\(\mu\)J)\\  \midrule
      SpiNN & 1000    & 7.1                     \\
      Loihi & 296  & 37                            \\
      
      \bottomrule

    \end{tabular}
  \end{minipage}
\end{table}


\subsection{Adaptive Control}\label{result_adaptive_control}

\subsubsection{Memory Footprint}\label{result_adaptive_control_memory}

For an adaptive control network simulated on a PE, the required SRAM memory mainly consists of 4 parts: input weight matrix and bias memory, output weight matrix memory, neuron input current memory and neuron memory.

The input weight matrix and bias memory is the memory for storing the input weight matrix and bias, which are quantized as 8-bit integers. The required memory in bytes is
\begin{equation}
M_{ib} \;=\; (D_{in} + 1) N
\label{eq:M_e-b}
\end{equation}

where \(D_{in}\) is the number of input dimensions, \(N\) is the number of neurons. 

The output weight matrix memory is the memory for storing the output weight matrix, which are 16 bit floating point numbers. The required memory in bytes is
\begin{equation}
M_{o} \;=\; 2D_{out} N
\label{eq:M_d}
\end{equation}

where \(D_{out}\) is the number of output dimensions. 

The neuron input current memory is the memory for storing the results from the MAC array after the input processing is complete. Each input current is a 32 bit integer. The required memory in bytes is
\begin{equation}
M_{ic} \;=\; 4N
\label{eq:M_ic}
\end{equation}

The neuron memory is the memory to hold the LIF neuron parameters like the membrane potential and refractory time. Each of them has 32 bits. The required memory in bytes is
\begin{equation}
M_{n} \;=\; 8N
\label{eq:M_n}
\end{equation}

The total memory for a neural network on a PE is
\begin{equation}
M_{total} \;=\; M_{ib} + M_{o} + M_{ic} + M_{n}
\label{eq:M_total}
\end{equation}

\begin{figure}[htb]
  \centering
  \includegraphics[width=0.48\textwidth]{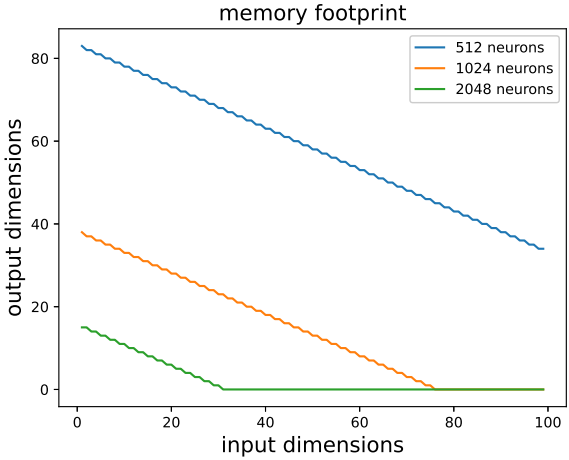}
  \caption{Maximum number of output dimensions for each input dimension and number of neurons for a neural network simulated on a PE.
  }\label{fig:adaptive_memory} 
\end{figure}

Since it is assumed that each PE has 90 KBytes of SRAM memory available for the data of the neural network, the maximum number of output dimensions given the number of input dimensions and number of neurons in a neural network can be derived with Eq. (\ref{eq:M_e-b}), (\ref{eq:M_d}), (\ref{eq:M_ic}), (\ref{eq:M_n}), (\ref{eq:M_total}). The result is shown Fig. \ref{fig:adaptive_memory}. 

\subsubsection{Computation Time and Comparison with Loihi}\label{result_adaptive_control_time}

\begin{figure}[htb]
  \centering
  \includegraphics[width=0.48\textwidth]{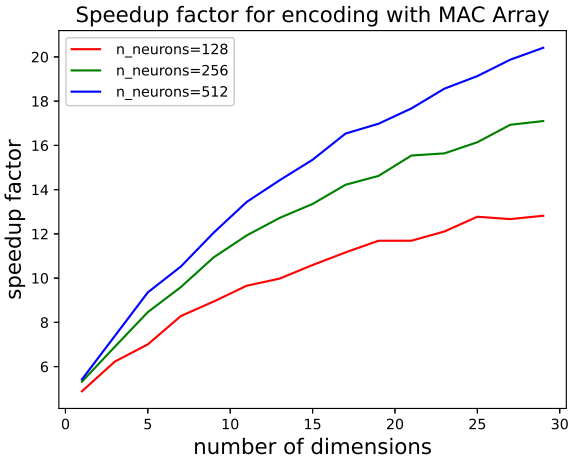}
  \caption{Speedup of input processing time with the MAC array
  }\label{fig:encoding2} 
\end{figure}

\begin{figure*}[ht]
  \centering
  \includegraphics[width=1.0\textwidth]{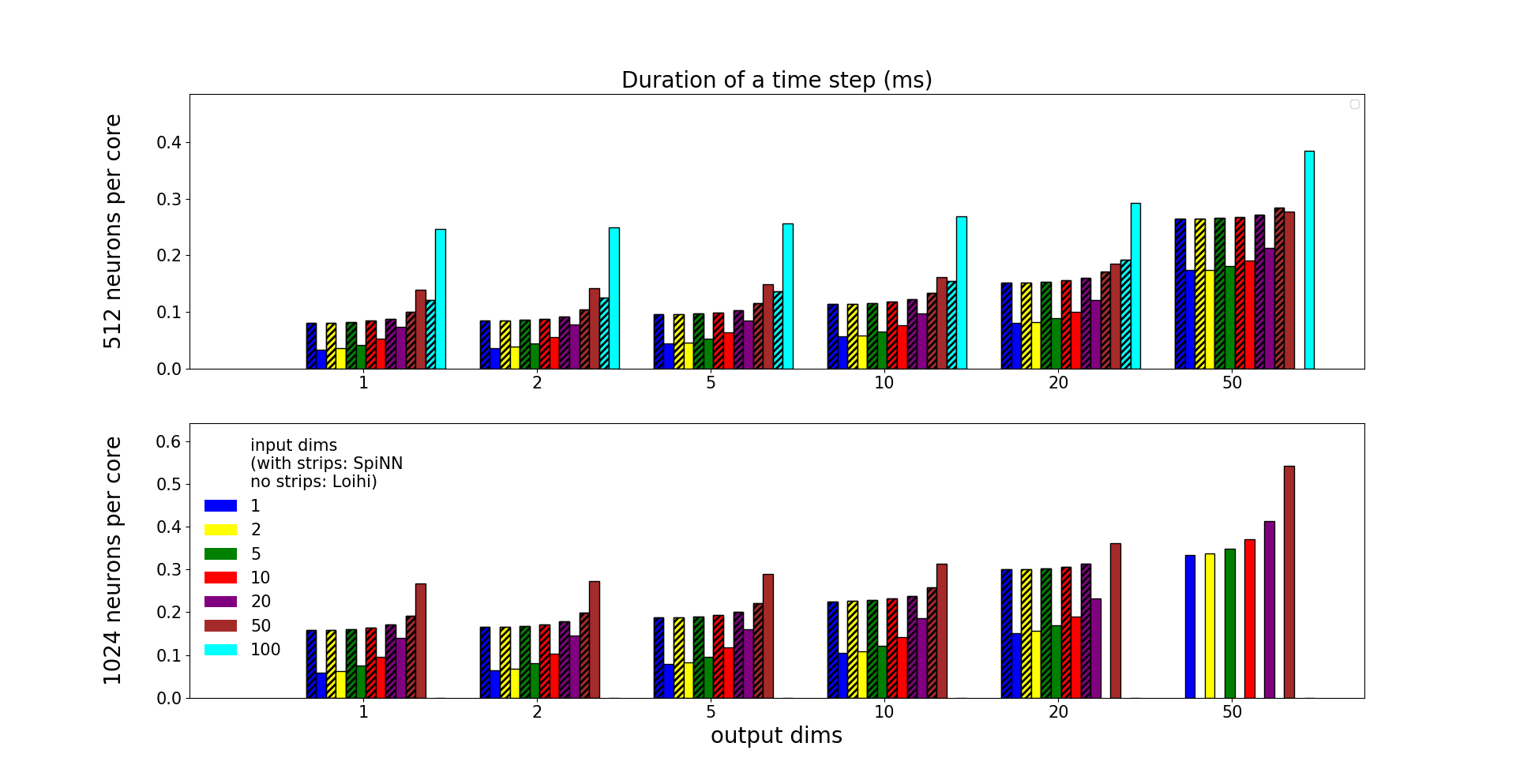}
  \caption{Duration of a time step of SpiNNaker 2 prototype (with strips) and Loihi (without strips) for different number of neurons per core, different input and output dimensions for the adaptive control benchmark.  No measurement result for the SpiNNaker 2 prototype is shown where the implementation is limited by memory.
  }\label{fig:loihi_jib1_comparison_time} 
\end{figure*}

For adaptive control, the computation times for input processing (\(T_{i\_mlacc}\) / \(T_{i\_no\_mlacc}\)), neuron update (\(T_{n}\)), output processing (\(T_{o}\)) and weight update (\(T_{w}\)) are measured. After the measurement, polynomial models can be fitted by minimizing the mean-squared error. For input processing with MAC array, the number of clock cycles is 
\begin{align}
T_{i\_mac} \;=\; 131.21  + 5.07 N \nonumber \\ 
+ 0.13 N D_{in} + 35.79 D_{in}
\label{eq:c_encoding_mac}
\end{align}

where \(N\) is the number of neurons, \(D_{in}\) is the number of input dimensions. Eq. (\ref{eq:c_encoding_mac}) is very similar to Eq. (\ref{eq:c_mm}), because the main computation is in both cases done by the MAC array. The difference is caused by the different data types. In keyword spotting, the inputs are assumed to be 8 bit integers, but in adaptive control, each input is assumed to be floating point. This is necessary because in general, the same implementation can be used as a building block for NEF implementation on SpiNNaker 2 to construct large-scale cognitive models as mentioned in Section \ref{adaptive_control}, so that the input data type needs to be the same as the output data type. Since the output weights are floating point, and their values change dynamically due to learning, an extra range check is performed for each input value, and an extra data type conversion is performed. This is reflected in \(35.79 D_{in}\) and the constant \(131.21\). 

The number of clock cycles without MAC array is 
\begin{align}
T_{i\_no\_mac} \;=\; 102.52  + 22.54  N \nonumber \\
 + 7.07  N  D_{in} + 25.54  D_{in}
\label{eq:c_encoding_no_mac}
\end{align}

The main benefit of MAC array is reflected in the reduction of \(7.07  N  D_{in}\) in Eq. (\ref{eq:c_encoding_no_mac}) to \(0.13 N D_{in}\) in Eq. (\ref{eq:c_encoding_mac}), which is made possible by the SIMD operation of the MAC array. The speedup is higher for higher dimensions. Fig. \ref{fig:encoding2} shows the speedup of the computation time for input processing with the MAC array compared to without the MAC array.

Unlike in keyword spotting, where the ReLU neuron model is used, in adaptive control, the LIF neuron model is used, which is the same as in Loihi. The neuron update time in terms of number of clock cycles is 
\begin{align}
T_{n} \;=\; 28.19 N  - 26.90 NP  + 509.18
\label{eq:c_lif_neuron_update}
\end{align}

where \(P\) is the firing probability. The minus sign in \(-26.9 N P\) is because during the refractory period, the computation needed is reduced. Since this is event based, it depends on \(P\). 

The output processing time is
\begin{equation}
T_{o} \;=\; 5.8  ND_{out}P + 19.31 NP
\label{eq:c_lif_decoding}
\end{equation}

where \(D_{out}\) is the number of output dimensions.

The weight update time is 
\begin{equation}
T_{w} \;=\; 8.28 ND_{out}P + 28.04N P
\label{eq:c_lif_learning}
\end{equation}

The total time is
\begin{align}
T_{total} \;=\; T_{i\_mac} + T_{n} + T_{o} + T_{w}
\label{eq:c_lif_total}
\end{align}

Since output processing and weight update are event based, the firing rate of 130 Hz corresponding to a firing probability \(P\) of 0.13, which is used for comparing the SpiNNaker 2 prototype with Loihi, would reduce the computation time by 87\% compared to a non-event-based implementation.

Typically, the SpiNNaker system runs in real time with 1 ms time step. When the PE is running at 250 MHz, the available number of clock cycles for each time step is 250 000, which is the computational constraint. According to Eq. (\ref{eq:c_lif_total}), for the range of the parameters shown in Fig. \ref{fig:adaptive_memory}, the computation can be done within 1 ms. So the maximum implementable size of a network on a single PE in this benchmark is constrained by memory rather than computation. 

Although the SpiNNaker system runs with a constant timer tick, and within a timer tick, the Arm core enters sleep mode after the computation is done, the study of the computation time provides information about how much the time step can be potentially reduced, e.g. when instead of a 1 ms timer tick, a 0.5 ms timer tick is required. 

For the adaptive control benchmark task with different number of input dimensions, output dimensions and number of neurons, the duration of a time step of SpiNNaker 2 prototype and Loihi is compared and shown in Fig. \ref{fig:loihi_jib1_comparison_time}, with the mean population firing rate kept at around 130 Hz for both hardwares. Here the duration of a time step for the SpiNNaker 2 prototype refers to the time for the PE to complete the computation of a time step. From the comparison it is clear that for small number of input dimensions, Loihi is faster than the SpiNNaker 2 prototype, and for large number of input dimensions, the SpiNNaker 2 prototype is faster than Loihi. The maximum ratio of duration of a time step between both hardwares is summarized in Tab. \ref{tab:adaptive_control_relative_time}.  

Because of the MAC array, the computation time of the SpiNNaker 2 prototype increases less rapidly with the number of input dimensions, so that the SpiNNaker 2 prototype could catch up with Loihi in terms of computation time for higher input dimensions. 

\begin{table}[htb]
  \begin{minipage}{0.47\textwidth}
    \centering
    \renewcommand{\arraystretch}{1.1}
    \caption{Maximum ratio of duration of a time step between the SpiNNaker 2 prototype (SpiNN) and Loihi for the adaptive control task }
    \label{tab:adaptive_control_relative_time}
    \centering
    \footnotesize
    \begin{tabular}{lcc} \toprule
      Input Dimensions &  1 & 100     \\  
       Output Dimensions & 1 & 1                      \\
       Number of Neurons &  1024 &  512                         \\
      Duration of a Time Step SpiNN : Loihi & 1 : 0.37 & 0.49 : 1\\
      \bottomrule

    \end{tabular}
  \end{minipage}
\end{table}


\subsubsection{Energy Measurement and Comparison with Loihi}\label{result_adaptive_control_power}

\begin{figure*}[ht]
  \centering
  \includegraphics[width=1.0\textwidth]{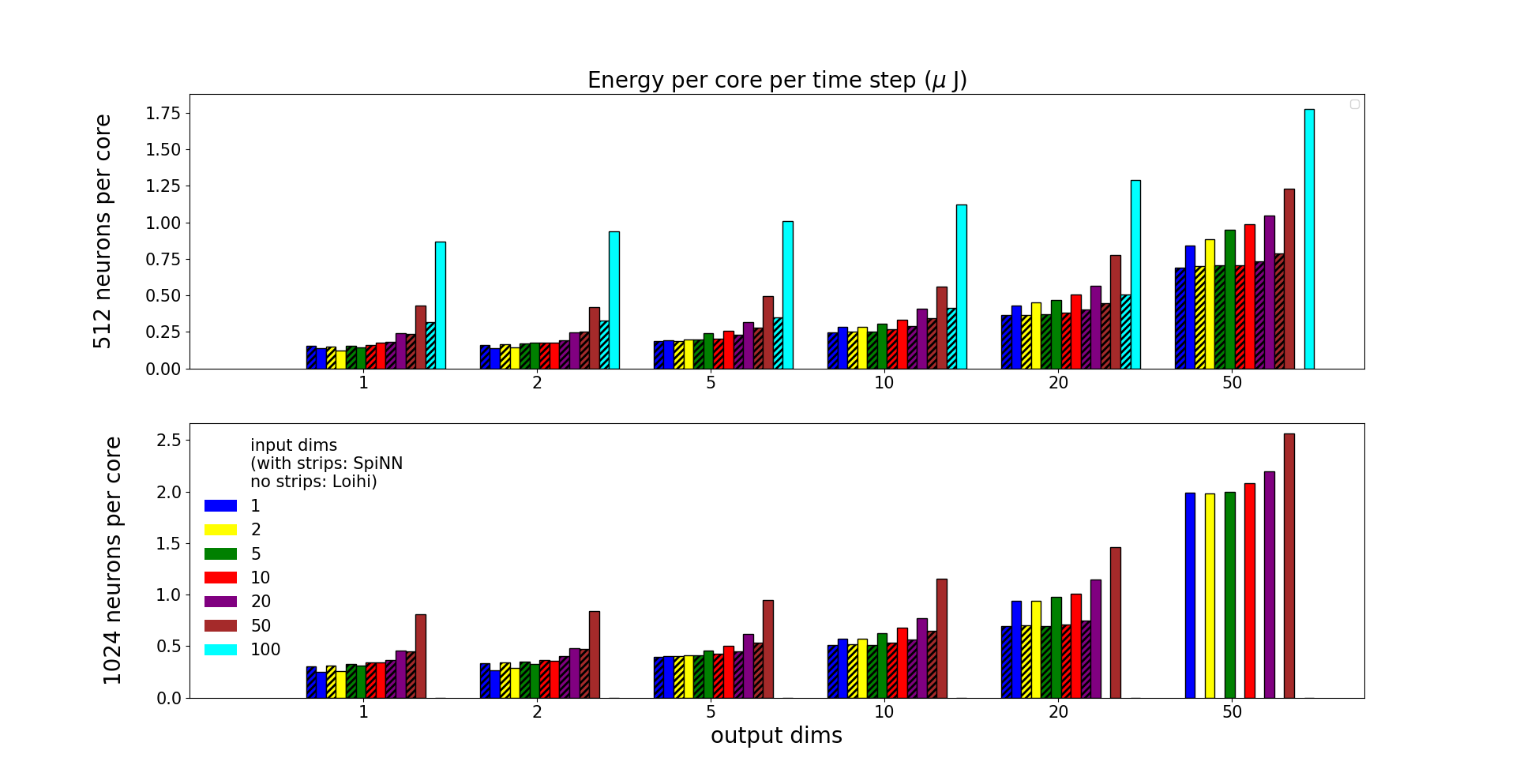}
  \caption{Active energy of SpiNNaker 2 prototype (with strips) and Loihi (without strips) for different number of neurons per core, different input and output dimensions for the adaptive control benchmark.  No measurement result for the SpiNNaker 2 prototype is shown where the implementation is limited by memory.
  }\label{fig:loihi_jib1_comparison} 
\end{figure*}

\begin{figure}[htb]
  \centering
  \includegraphics[width=0.48\textwidth]{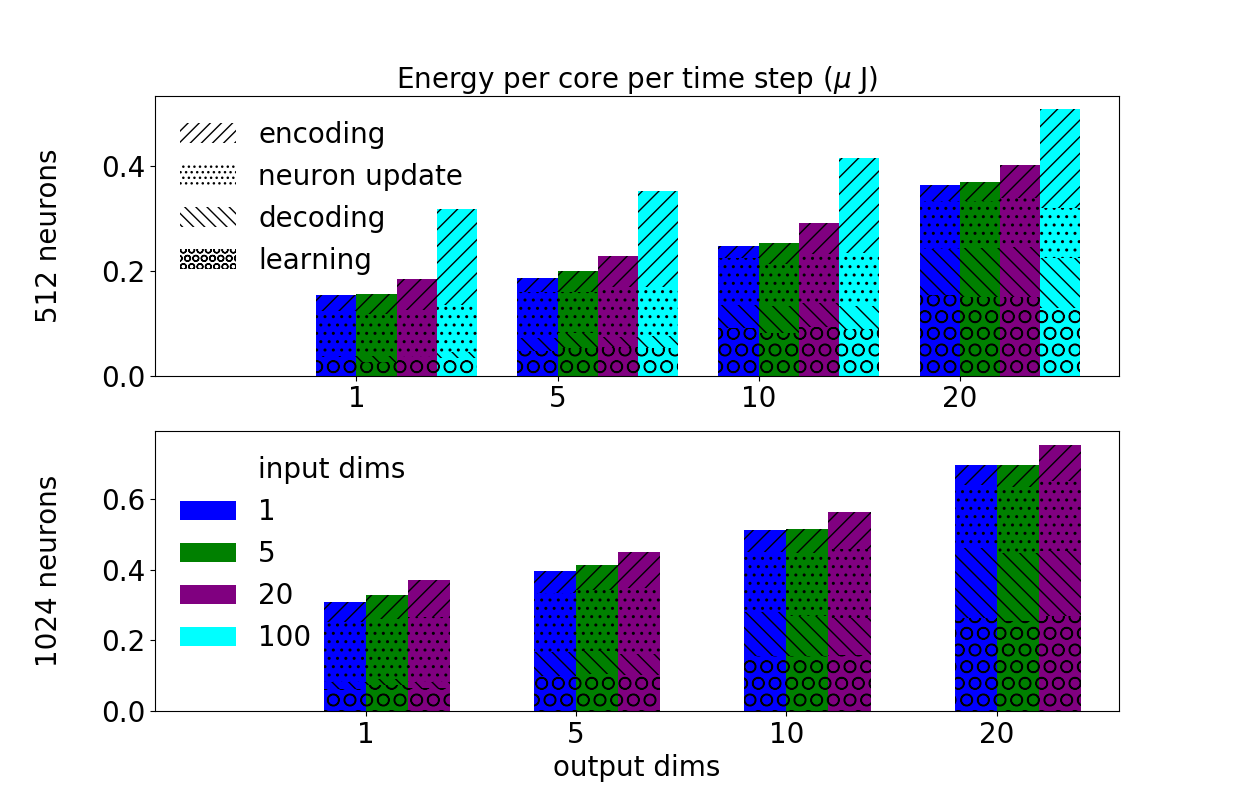}
  \caption{Breakdown of energy consumption per core per time step of the SpiNNaker 2 prototype into 4 energy components: input processing, neuron update, output processing and weight update. 
  }\label{fig:jib1_energy_breakdown} 
\end{figure}


The energy consumption of the SpiNNaker 2 prototype and Loihi is measured with the same parameters as in the computation time comparison. The result is shown in Fig. \ref{fig:loihi_jib1_comparison}. Similar to Section \ref{result_keyword_spotting_power}, only the active energy is shown. For small number of input dimensions, Loihi is more energy efficient than the SpiNNaker 2 prototype, and for large number of input dimensions, the SpiNNaker 2 prototype is more energy efficient than Loihi. The maximum ratio of active energy consumption between both hardwares is summarized in Tab. \ref{tab:adaptive_control_relative_energy}.

\begin{table}[htb]
  \begin{minipage}{0.47\textwidth}
    \centering
    \renewcommand{\arraystretch}{1.1}
    \caption{Maximum ratio of active energy consumption between the SpiNNaker 2 prototype (SpiNN) and Loihi for the adaptive control task }
    \label{tab:adaptive_control_relative_energy}
    \centering
    \footnotesize
    \begin{tabular}{lcc} \toprule
      
      Input Dimensions &  1 & 100     \\  
       Output Dimensions & 1 & 1                      \\
       Number of Neurons &  1024 &  512                         \\
      Active Energy SpiNN : Loihi & 1 : 0.81 & 0.36 : 1 \\
      \bottomrule

    \end{tabular}
  \end{minipage}
\end{table}

Similar to the computation time comparison, we see the benefit of MAC array especially for high input dimensions, when the MAC array is more extensively used. This is made more clear in the energy breakdown in Fig. \ref{fig:jib1_energy_breakdown}. Here, it is clear how the input processing energy increases with the input dimensions for the same number of neurons and output dimensions, how the neuron update energy increases with the number of neurons for the same input dimensions and output dimensions, and how the output processing and weight update energy increases with the number of output dimensions for the same input dimensions and number of neurons.


\subsubsection{Robotic Demo}\label{result_adaptive_control_demo}

The SpiNNaker 2 prototype running the adaptive control benchmark is connected to a robotic arm built with Lego Mindstorms EV3 robot kit. The setup is based on \cite{terry15}. The input to the neural network is the position and velocity of the motor and the output of the neural network is the motor control signal to be combined with the PD controller output, as described in Section \ref{adaptive_control}. 

\begin{figure}[h]
  \centering
  \includegraphics[width=0.48\textwidth]{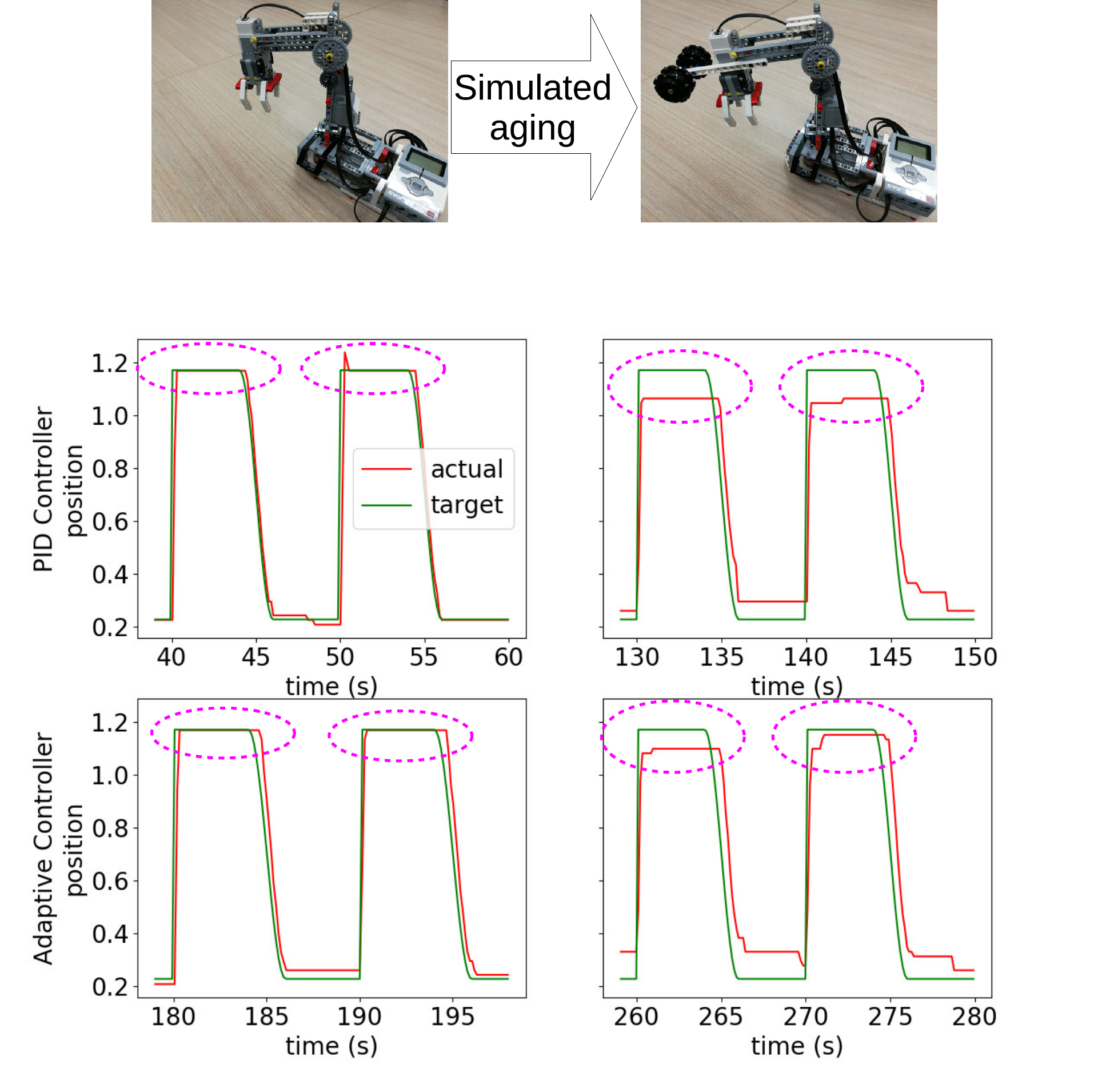}
  \caption{Robotic demo. Upper part: In the normal case (left), there is no extra weight attached to the robotic arm. In the simulated aging case (right), an extra weight is attached to resemble the aging effect. Lower part: Performance of the PID Controller and the adaptive controller in the normal case and the simulated aging case. The red curve shows the actual position of the arm, and the green curve shows the target position. The shown position is the normalized angle position of the motor. In the normal case, both the PID Controller and the adaptive controller perform well. But in the simulated aging case, the PID Controller cannot adapt to the new situation, while the adaptive controller can adapt to the new situation by learning from error feedback, thus improving the performance, as seen in the improvement from the first to second motor commands.
  }\label{fig:robotic_demo1} 
\end{figure}

In this demo we consider two situations: the normal case and the simulated aging case (Fig. \ref{fig:robotic_demo1}, upper part). In the case of simulated aging an extra weight is added to the robotic arm to resemble the aging effect or unknown disturbance. For each case the performance of the adaptive controller is compared with a normal PID controller. In the normal case, both controllers perform equally well, but in the simulated aging case, the PID controller cannot adapt itself to the new situation, while the adaptive controller can learn from the error feedback and adapt its parameters to improve the performance (Fig. \ref{fig:robotic_demo1}, lower part).

\section{Discussion}
\label{sec:discussion}

In this section, we consider the suitability of other neuromorphic platforms for implementing the benchmarks in this work. Since the comparison between the SpiNNaker 2 prototype and Loihi has already been extensively discussed in previous sections, we leave the summary of this comparison to the Conclusion section. 

\subsection{Comparison with SpiNNaker 1}

We assume that the same benchmarks in this work could also be implemented on SpiNNaker 1. However, since in SpiNNaker 1 there is no MAC array, the vector-matrix multiplication would be much slower and therefore consume much more energy than the SpiNNaker 2 prototype. Fig. \ref{fig:encoding2} indicates the speedup in terms of number of clock cycles for the vector-matrix multiplication in the SpiNNaker 2 prototype compared to what it would be in SpiNNaker 1. The differences in fabrication technology and supply voltage etc. further increases the difference between the SpiNNaker 2 prototype and SpiNNaker 1. 


\subsection{Comparison with other neuromorphic platforms}

To ease the discussion, we group neuromorphic platforms into 3 categories:

1. Neuromorphic platforms with static synapses, such as TrueNorth \cite{Merolla668}, NeuroGrid \cite{neurogrid}, Braindrop \cite{braindrop}, HiAER-IFAT \cite{park2017hierarchical}, DYNAPs \cite{dynaps18}, Tianjic \cite{tianjic}, NeuroSoC \cite{bionect20} and DeepSouth \cite{deepsouth18},

2. Neuromorphic platforms with configurable (but not programmable) plasticity, such as ROLLS \cite{rolls15}, ODIN \cite{odin19} and TITAN \cite{titan16},

3. Neuromorphic platforms with programmable plasticity, such as (except SpiNNaker 1/2 and Loihi) the BrainScales~1/2 system\cite{PPU,hartmann2010highly}.

We assume all 3 groups of neuromorphic platforms should be able to implement the keyword spotting benchmark in this work. However, DNNs can not be directly implemented on these platforms since they only support SNNs (except Tianjic, which also supports DNNs). Solutions similar to the SNN version implemented on Loihi would be an option. 

For adaptive control, since learning is involved, we assume the neuromorphic platforms in group 1 can not support this benchmark. It would be still possible to have an external host PC to reprogram the synaptic weights, but that would not be suitable for embedded applications. 

Although the learning rule in adaptive control is relatively simple, it involves multiplying an external error signal with the activity of the presynaptic neuron in every time step, which is quite different from the learning rules normally supported in the neuromorphic community, like Spike-Timing Dependent Plasticity (STDP) \cite{Markram213} or Spike-Driven Synaptic Plasticity (SDSP) \cite{brader07}. Therefore we assume the neuromorphic platforms in group 2 could not implement the adaptive control benchmark. 

The BrainScales~2 system in group 3 comes with programmable plasticity, but since the neural network runs in accelerated time, it is unclear whether the neural activity of each time step can be used for the weight update. Also it is unclear how to interface robotic applications which require real time response with a neural network running in accelerated time.

\section{Conclusion}
The PE of the SpiNNaker 2 prototype consists of a general purpose processor plus highly efficient accelerators, while Loihi employs dedicated circuits for neuron and synapse models plus a flexible learning engine. In this work, we compare these two platforms by comparing their performance in the same applications, namely keyword spotting and adaptive control. 

For keyword spotting, because of the MAC array used for vector-matrix multiplication and Arm core used for ReLU activation, the DNN version of keyword spotting network can be directly implemented on the SpiNNaker 2 prototype, while on Loihi the SNN version is implemented for the same task. The result of this is faster inference and higher energy efficiency of the SpiNNaker 2 prototype. 

For adaptive control both the SpiNNaker 2 prototype and Loihi are efficient in specific parameter regions. The SpiNNaker 2 prototype is more efficient than Loihi both regarding the computation time and active energy, when the number of input dimensions is high, because that is where the vector-matrix multiplication is more complicated and the MAC array is more dominant. On the other hand, the SpiNNaker 2 prototype is less efficient than Loihi when the number of input dimensions is low, because that is where the vector-matrix multiplication is less complicated and the Arm core is more dominant. 

Through the comparison of the SpiNNaker 2 prototype and Loihi in these two benchmarks, we try to bring more insight into the SpiNNaker 2 system and highlight the benefit of the MAC array in neuromorphic applications. Since both SpiNNaker 2 and Loihi have very wide application fields, the two benchmarks in this work is by far not a comprehensive comparison of both neuromorphic platforms. The comparison regarding other benchmarks would be out of scope of this work and is left for future work. 


\bibliographystyle{IEEEtran}
\bibliography{nengo_robotics}

\clearpage


\end{document}